\documentclass[runningheads]{llncs}

\usepackage{eccv}

\usepackage{eccvabbrv}

\usepackage{graphicx}
\usepackage{booktabs}

\usepackage[accsupp]{axessibility}  % Improves PDF readability for those with disabilities.

\usepackage{hyperref}
\usepackage{sidecap}
\usepackage{floatrow}
\usepackage{algorithm}
\usepackage{algpseudocode}
\usepackage{enumitem}

\usepackage{wrapfig,booktabs}

\usepackage{bm}

\usepackage{xr}

\usepackage{orcidlink}

\newcommand{\branch}{\beta}
\newcommand{\threeDTree}{T}
\newcommand{\fourDTree}{H}

\newcommand{\srvf}{q}
\newcommand{\pcasrvf}{w}
\newcommand{\srvfbranch}{\srvf}
\newcommand{\srvfthreeDTree}{Q}
\newcommand{\srvffourDTree}{\srvf} %D}

\newcommand{\preshapespace}{\mathcal{C}}
\newcommand{\preshapespaceThreeDTreees}{\preshapespace_\threeDTree}
\newcommand{\preshapesrvts}{\preshapespace_\srvfthreeDTree}

\newcommand{\pcaspace}{\preshapespace_{\text{PCA}}}
\newcommand{\srvfpcaspace}{\preshapespace_\pcasrvf}

\newcommand{\rotation}{O}
\newcommand{\rotationspace}{SO(3)}
\newcommand{\reparm}{\gamma}
\newcommand{\diffeotree}{\boldsymbol{\reparm}}
\newcommand{\reparmspace}{\Gamma}
\newcommand{\permutation}{\sigma}

\newcommand{\permutetree}{\boldsymbol{\permutation}}

\newcommand{\geodesic}{\alpha}

\newcommand{\fourDcurve}{\alpha}
\newcommand{\preshapespaceFourDcurve}{\preshapespace_\fourDcurve}
\newcommand{\fourDcurveinpca}{\alpha^{\text{pca}}}
\newcommand{\fourDcurveinpcaone}{\fourDcurveinpca_1}
\newcommand{\fourDcurveinpcatwo}{\fourDcurveinpca_2}

\newcommand{\covMatrix}{C}

\newcommand{\reparmcurve}{\xi}   %% don't use xi because i is usually used for indices  RESPONSE: it represents epsilon sign 
\newcommand{\reparmcurvespace}{\Xi} %% same as above

\newcommand{\optimal}{^*}

\newcommand{\geodbetnfourDtrees}{\Lambda}

\newcommand{\totalthreeDtree}{m}
\newcommand{\totalfourDtree}{n}

\newcommand{\eigenvectorcurvepca}{\delta}
\newcommand{\meantree}{\boldsymbol{\mu}}
\newcommand{\meansrvft}{\boldsymbol{\mu}_{\srvfbranch}}

\def\argmin{\mathop{\rm argmin}}

\renewcommand{\ie}{\emph{i.e., }}
\renewcommand{\eg}{\emph{e.g., }}

\newcommand{\ltwo}{\mathbb{L}^2}
\newcommand{\real}{\mathbb{R}}
\newcommand{\rthree}{\real^3}
\newcommand{\noi}{\noindent}

\begin{document}

\title{A Riemannian Approach for Spatiotemporal Analysis and Generation of 4D Tree-shaped Structures} 

\titlerunning{Spatiotemporal Analysis and Generation of 4D Tree-shaped Structures}

\author{Tahmina Khanam\inst{1}\orcidlink{0000-0002-9549-2700} \and
Hamid Laga\inst{1}\orcidlink{0000-0002-4758-7510} \and
Mohammed Bennamoun\inst{2}\orcidlink{0000-0002-6603-3257} \and
Guanjin Wang\inst{1}\orcidlink{0000-0002-5258-0532} \and
Ferdous Sohel\inst{1}\orcidlink{0000-0003-1557-4907} \and
Farid Boussaid\inst{2}\orcidlink{0000-0001-7250-7407} \and
Guan Wang\inst{3}\orcidlink{0000-0003-3029-0996} \and
Anuj Srivastava\inst{4}\orcidlink{0000-0001-7406-0338}}

\authorrunning{K.~Tahmina et al.}

\institute{Murdoch University, WA, Australia 
\email{34719017@student.murdoch.edu.au, \{H.Laga, Guanjin.Wang,F.Sohel\}@murdoch.edu.au} \and
University of Western Australia, WA, Australia  \email{\{mohammed.bennamoun, farid.boussaid\}@uwa.edu.au} \and Yangtze Delta Region Institute (Huzhou), University of Electronic Science and Technology of China, China \email{wangguan12621@gmail.com} \and Florida State University, USA  \email{anuj@stat.fsu.edu}}

\maketitle

\begin{abstract}
  We propose the first comprehensive approach for modeling and analyzing the spatiotemporal shape variability in tree-like 4D objects, \ie  3D objects whose shapes bend, stretch and change in their branching structure over time as they deform, grow, and interact with their environment.  Our key contribution is the representation of tree-like 3D shapes using Square Root Velocity Function Trees (SRVFT) \cite{Guan_tree}. By solving the spatial registration  in the SRVFT space, which is equipped with an $\ltwo$ metric, 4D tree-shaped structures become time-parameterized trajectories in this space. This reduces the problem of modeling and analyzing 4D tree-like shapes to that of modeling and analyzing elastic trajectories in the SRVFT space, where elasticity refers to time warping. In this paper, we propose a novel mathematical representation of the shape space of such trajectories, a Riemannian metric on that space, and computational tools for fast and accurate spatiotemporal registration and geodesics computation between 4D tree-shaped structures. Leveraging these building blocks, we develop a full framework for modelling the spatiotemporal variability using statistical models and generating novel 4D tree-like structures from a set of exemplars. We demonstrate and validate the proposed framework using real 4D plant data. %The code is available on Github \url{https://github.com/Tahmina979/4Dtreeshape_project}  
\end{abstract}

%% ======= Introduction
% Here is the short version
\section{Introduction}
We propose a novel framework for the statistical analysis of the spatiotemporal shape variability in tree-like 3D objects that deform over time, hereinafter referred to as 4D (or 3D $+$ time) tree-like structures. Such objects, which undergo complex non-rigid and topological deformations, are abundant in nature. Understanding and modelling the  deformation patterns of their shapes can benefit many applications, from plant biology and medicine to 4D content generation in computer graphics and virtual reality. For example, plants  and blood vessels deform over time due to either normal growth or the progression of diseases. Modeling the spatiotemporal variability can help distinguish shape deformations that are due to normal growth from those due to disease progression.

This problem has received growing attention. However, most of the work focused on the 4D reconstruction of dynamic objects \cite{jiang2022lord,jiang2022h4d} and on the statistical analysis of 4D shapes that bend and stretch while preserving their topology \cite{4D_atlas}. In contrast, this paper focuses, for the first time, on 4D shapes that bend, stretch, and change in topology. Given a set of such 4D objects,  our goal is to learn statistical summaries, such as the mean and modes of variation, characterize the spatiotemporal shape variability in the set using statistical models, and learn a generative model capable of synthesizing and generating novel 4D tree-shaped structures. Achieving these goals requires accurate spatiotemporal registration of the 4D tree-like shapes. Spatial registration refers to the problem of finding one-to-one branchwise and pointwise correspondences across 3D shapes. This is very challenging in our setup since tree-like 3D shapes such as plants and botanical trees exhibit not only large elastic deformations, \ie the bending and stretching of their branches,  but also topological differences, \ie differences in the branching structures,  within and across shape classes. Temporal registration refers to the problem of temporally aligning 4D shapes. For example, two botanical plants can grow at different rates even if they are of the same species. This adds another level of complexity, which is the high temporal variability due to differences in the execution rates (speeds) between 4D tree-shaped objects.

\begin{figure}[t]
    \centering
    \includegraphics[width=\linewidth]{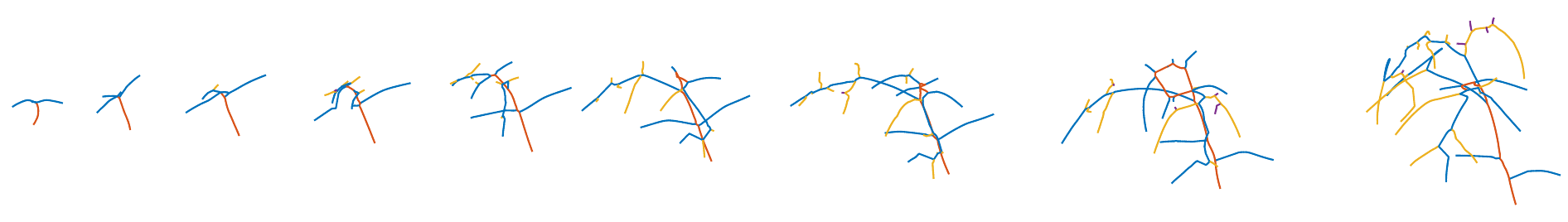}
    \caption{The 4D skeleton of Fig. 1 in the Supplementary Material, structured it into layers of branches. \textcolor{red}{The red branch is the main branch}, \textcolor{blue}{blue corresponds to the $2^{nd}$ layer of branches}, \textcolor{yellow}{yellow to the $3^{rd}$ layer}, and \textcolor{purple}{purple to  the $4^{th}$ layer}.}
    \label{fig:real_tree_vis}
\end{figure}

Our key contribution is the representation of tree-like 3D objects using Square Root Velocity Functions (SRVF) \cite{srivastava2010shape}. By solving the spatial registration problem in the SRVF space, tree-like 4D shapes become trajectories in this space. This reduces the problem of modeling and analyzing 4D tree-like shapes to that of modeling and analyzing elastic trajectories in the SRVF space, where elasticity refers to time warping. Building on \cite{Guan_tree}, which developed tools for the spatial registration of tree-like 3D shapes in the extended SRVF space, we propose \textbf{(1)} a set of computational tools for the spatiotemporal registration of 4D tree-like shapes, \textbf{(2)}  a mechanism for computing geodesics between 4D trees,  \textbf{(3)} a framework for the statistical analysis of the spatiotemporal shape variability, and \textbf{(4)} a mechanism for learning, from a set of exemplars, generative models capable of synthesizing novel 4D tree-like structures. We demonstrate the efficiency of the proposed tools using real 4D shapes of growing tomato and maize plants.    

The  paper is organized as follows. \cref{sec:related_work} summarizes the related work.  \cref{sec:3D_shape_space} presents the proposed SRVF representation of tree-like 3D shapes.   \cref{sec:4D_tree_shape_space} presents the proposed computational tools for the spatiotemporal registration and  geodesics computation between 4D tree-shaped structures.  \cref{sec:statistical_analysis} demonstrates how these tools can be used to compute summary statistics  and synthesize novel 4D tree-shapes. \cref{sec:result} presents the results while \cref{sec:conclusion} concludes the paper.

%== Related work
\section{Related Work}
\label{sec:related_work}

We refer to~\cite{laga2018survey,4D_atlas,Guan_tree} for a detailed  survey. Here, we focus on 4D shape analysis and related topics. 
Studying and modelling the spatio-temporal shape variability in 4D shapes  requires defining a proper shape space and metrics on this space, performing spatiotemporal registration under the metric, and computing geodesics between 4D shapes.  The bulk of the work is focused on the spatial registration problem, which is traditionally solved using hand-crafted feature matching and the Iterated Closest Point~\cite{besl1992method,zhang2021fast,tsumura2023body}. These methods are well suited for 3D objects that bend and stretch, \eg human/animal body shapes. Tree-like 3D objects such as plants  and blood vessels have complex structures and exhibit topological variability, which makes their spatial registration challenging. 

To address these issues, Feragen \etal~\cite{simpletree1,simpletree2,simpletree3,simpletree4}  treat a tree-shaped object as a point in a tree shape space equipped with the Quotient Euclidean Distance (QED) as a  metric. The problem of registering two tree-shaped objects is then reduced to that of finding the optimal deformation path, or geodesic with respect to the metric, that deforms one tree onto the other. These deformations include branch bending, stretching, and topological changes. Under the QED, bending and stretching are measured using the $\ltwo$ metric while topological changes are measured using graph edit distance. This results in a significant shrinkage along the geodesic paths when the deformations are large. Wang \etal~\cite{QED,wang2018statistical} alleviated this issue by representing the shape of the edges in the tree shapes of Feragen \etal~\cite{simpletree1,simpletree2,simpletree3,simpletree4}  using the Square Root Velocity Functions (SRVF)~\cite{srivastava2010shape}  and then using the QED in the SRVF space, which is equivalent to the full elastic metric in the original space. This still suffers from the shrinkage that is due to the edge collapses used by the QED metric.  Duncan \etal~\cite{duncan_bifurcation}  addressed these issues  by representing trees using sliding branches, instead of collapsing edges. Guan \etal~\cite{Guan_tree}  extended this representation to tree-like 3D shapes such as complex botanical trees, plant roots, and neuronal structures. It has also been used to build a full generative model of tree-like 3D shapes by fitting Gaussian distributions to tree-like 3D shapes represented as points in the tree-shape space.

This paper builds upon and generalizes the representation of~\cite{Guan_tree} to 4D tree-shaped objects, \ie sequences of tree-like 3D shapes that deform time. The key idea is to represent 4D shapes as trajectories in the tree-shape space reducing the problem to that of analyzing elastic trajectories in the tree-shape space. This idea has been introduced by Laga \etal~\cite{4D_atlas}  for objects that have a manifold structure and only bend and stretch, but do not change in topology. This paper focuses, for the first time, on 3D shapes that also change in topology. Several recent papers in plant biology focused on the registration of 4D plants~\cite{magistri2020segmentation,chebrolu2020spatio,skeletal_extraction,chebrolu2021registration,wang2022plantmove,zhang2023spatio,lobefaro2023estimating}. They, however, define registration of 4D plants as the spatial registration of 3D plants within plant sequences. This is spatial registration. In this paper, we focus, for the first time, on spatial and temporal registration and develop a comprehensive suite of computational tools that enable the computation of geodesics, summary statistics, and generative models for 4D tree-shaped objects.

%% 3D tree space
\section{The Space of 3D tree-shaped structures}
\label{sec:3D_shape_space}
We first present the mathematical framework we use to represent tree-structured 3D shapes (\cref{sec:representation}), the elastic metric for quantifying their bending, stretching, and topological changes (\cref{sect:srvf_3Dtree}), and the computational tools for their spatial registration and geodesics computation  (\cref{sect:spatial_registraion_within}). Throughout the paper, we  use the Pheno4D dataset~\cite{pheno4d}, which consists of multiple 4D plants. Each 4D plant is a sequence of 3D plants, in the form of point clouds, captured at different  times during its growth. We extract the skeleton of the plants using~\cite{skeletal_extraction} and partition it into branches; see~\cref{fig:real_tree_vis} and also Fig. 1 in the Supplementary Material.

\subsection{Representation}
\label{sec:representation}
We represent a branch of a 3D tree structure $\threeDTree$ as an arc length-parameterized curve $\branch : [0, 1]  \rightarrow \rthree$ where $\branch(s) = (x(s), y(s), z(s))$. We structure a 3D tree into layers; see \cref{fig:real_tree_vis}. The first layer $\branch^0$ is the main branch with $k^0$ sub-trees $\threeDTree^1_i$ attached to it at the bifurcation points $s^0_i, i=1,\dots,  k^0$. That is, $
    \threeDTree = (\branch^0, \{\threeDTree^1_{i}, s^0_{i}\}_{i=1}^{k^0}) 
$.  A 3D subtree $\threeDTree^l$ at level $l$ of the hierarchy can be represented recursively as $
 \threeDTree^l = (\branch^l,\{\threeDTree^{l+1}_i,s^{l}_i\}_{i=1}^{k^l}), \text{ for } l = 1, \dots, L-1$. Here, $L$ is the total number of layers,  $\branch^l$ is the main branch of  $\threeDTree^l$, $k^l$ is the number of subtrees attached to the branch $\branch^l$, and $s^{l}_i \in [0, 1]$ is a bifurcation point on $\branch^l$ of  $\threeDTree^l$. The branches at different levels can have any arbitrary number of subtrees attached to them.  
A good representation needs to be invariant to similarity-preserving transformations (translation, scale, rotation, reparameterization). To achieve translation invariance, we translate each tree so that the start point of its main branch is located at the origin. Invariance to scale is optional as it may not be required for some applications such as growth analysis. The space of all such normalized trees is referred to as the \emph{pre-tree-shape space} and is denoted by $\preshapespace_{\threeDTree}$. The invariance to rotation and reparameterization will be dealt with at the metric level (\cref{sect:srvf_3Dtree}). With this representation, a 3D tree becomes a point in the pre-tree shape space $\preshapespaceThreeDTreees$. A   4D tree $\fourDTree$ can then be seen as a time-parameterized curve $\fourDTree: [0, 1] \to \preshapespaceThreeDTreees$ where $[0, 1] $ is the time domain.

\begin{figure}[t]%{r}{0.5\textwidth}
    \vspace{-0pt}
    \centering
    \includegraphics[width=\linewidth]{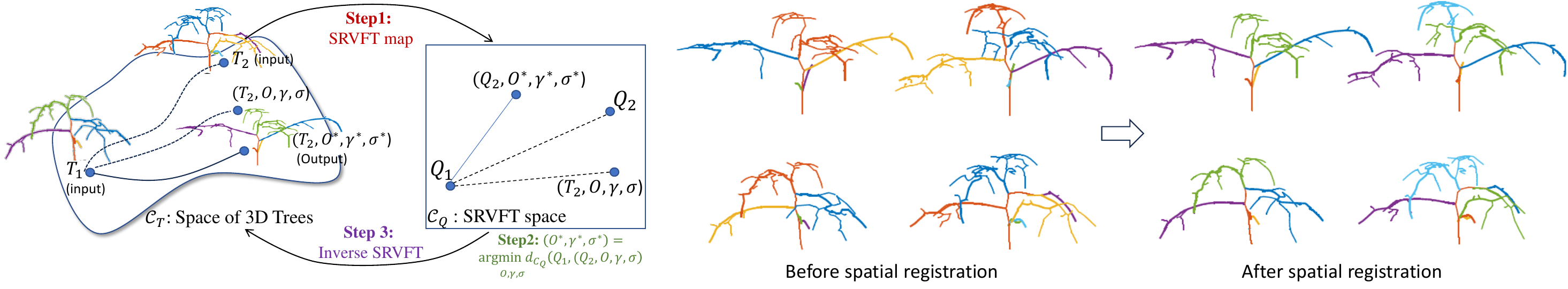}
    \caption{The proposed spatial registration. We also show four  3D tree shapes before and after their spatial registration using the proposed framework. Our  spatial registration took on average $235$s to align two tomato and $17$s to align two maize 4D plants.}
    \label{fig:spatial registration}
\end{figure}

\subsection{The elastic metric for comparing tree-shaped structures}
\label{sect:srvf_3Dtree}
If two 3D tree structures have the same number of branches, then one can quantify the dissimilarity between the two trees by measuring the amount of bending,  stretching, and branch sliding that one needs to apply to the branches of one tree  to align it onto the other.  
However, instead of explicitly using these metrics, which are non-linear and thus computationally expensive to evaluate and optimize, we use the  SRVF representation~\cite{srivastava2010shape} of 3D curves. Mathematically, the SRVF  $\srvfbranch$ of a curve $\branch$ is defined as:
\begin{equation}
\label{eq:SRVF}
    \small{
    \srvfbranch(\branch)(s) = 
         \frac{\branch'(s)}{\sqrt{||\branch'(s)||}},     \text{ if } \|\branch'(s)\|\neq 0, \text{ and } 0         \text{ otherwise}.
         }
\end{equation}

\noi Its main property is that the $\ltwo$ metric in the SRVF space is equivalent to a weighted sum of bending and stretching in the original space. It is also invertible, up to translation. In other words, given an SRVF, one can retrieve, up to translation, its corresponding original 3D curve. This significantly simplifies the analysis tasks: instead of measuring the similarity between two 3D curves with a complex elastic metric, one can map them to the SRVF space, perform the analysis there using the $\ltwo$ metric and then map the results back to the original space. 
Let  $\srvfthreeDTree$ be the SRVF Tree (SRVFT)  of an entire 3D tree $\threeDTree$ defined by computing the SRVF of each of its branches and appending their location with respect to their parent branch, and $\preshapesrvts$ the pre-shape space of SRVFTs. A proper metric on that space needs to be invariant to the global rotation $\rotation \in \rotationspace$, reparameterization $\diffeotree$ of the branches, and permutations $\permutetree$ of the orders of the lateral subtrees attached to a branch. We define $\diffeotree$ and $\permutetree$ recursively, \ie $\diffeotree = (\reparm^0, \{\diffeotree^{i}\}_{i=1}^{k^0})$ is the reparameterization of the main branch and its subtrees.  Specifically,  $\reparm^0 \in \reparmspace$ is a diffeomorphism that applies to the main branch of $\srvfbranch$ and $\diffeotree^{i}$ is the reparameterization, defined recursively, of the $i-$th subtree attached to the main branch. $\permutetree = (\permutation^0, \{ \permutetree^{i}\}_{i=1}^{k^0})$ where $\permutation^0 \in \permutetree$ is the permutation of the orders of the lateral subtrees on their corresponding main branch $\srvfbranch^0$, and $\permutetree^{i}$ defines recursively these permutations for the $i-$th subtree. Following~\cite{Guan_tree,duncan_bifurcation}, we define a rotation, reparameterization, and index permutation-invariant distance between two 3D trees, represented by their SRVFTs $\srvfthreeDTree_1$ and $\srvfthreeDTree_2$,  as the infimum over all possible rotations, branch reparameterizations, and branch index permutations:
\begin{equation}
	\label{eq:invariantdistance}
 \small{
	\begin{aligned}
		d\left(\srvfthreeDTree_1, \srvfthreeDTree_2\right) = \inf_{\tiny{\begin{tabular}{c} 
							 	$\rotation, \diffeotree , \permutetree $
							\end{tabular}
							}}  d_{\preshapesrvts}\left(\srvfthreeDTree_1, (\srvfthreeDTree_2,  \rotation, \diffeotree, \permutetree\right)), \text{ and }
	\end{aligned}
 }
\end{equation}\vspace{-12pt}
\begin{equation}
	\label{eq:invariant_metric}
 \resizebox{0.93\hsize}{!}{$\displaystyle{
	d_{\preshapesrvts}\left(\srvfthreeDTree_1, (\srvfthreeDTree_2,  \rotation, \diffeotree, \permutetree) \right)  = \lambda_m \parallel\srvfbranch^0_{1} - \rotation(\srvfbranch^0_{2}, \reparm_0) \parallel^2  + 				
				\lambda_p \sum_{i=1}^{n} \left({s}^{i}_{1} - {s}^{\permutetree(i)}_{2}\right)^2 
+ \lambda_s \sum_{i=1}^{n} d \left( \srvfthreeDTree_1^i, \srvfthreeDTree_2^{\permutetree(i)}   \right).}
	$}
\end{equation}

\noi Here,  $(\srvfthreeDTree, \rotation,  \diffeotree, \permutetree) $ denotes the result of applying these transformations to $\srvfthreeDTree$.

\subsection{Spatial registration and geodesics}
\label{sect:spatial_registraion_within}
\cref{fig:spatial registration} summarizes the spatial registration process.
With this representation, the optimal  rotation, diffeomorphism, and permutations that align $\srvfthreeDTree_2$ onto $\srvfthreeDTree_1$ can  be found by solving the following optimization problem; see Section 2 in the Supplementary Material for the detailed algorithm:
\begin{equation}
	\label{eq:registration}
	(\rotation\optimal,  \diffeotree\optimal, \permutetree\optimal) = \argmin_{\rotation, \diffeotree,\permutetree} d_{\preshapesrvts}\left(\srvfthreeDTree_1, (\srvfthreeDTree_2, \rotation,   \diffeotree, \permutetree) \right).
\end{equation}
\noi In practice, the input trees have different numbers of branches, thus they are elements of different subspaces. We address this  by adding null branches at the different levels of the trees. This way, trees become elements of the same pre-tree shape
space. The location of each additional null branch, \ie the
value of its parameter $s$, is initialized to the $s$ value of its
initial corresponding branch. These  will  get updated during the branchwise correspondences.

\cref{eq:registration}  registers onto each other a pair of tree-like 3D shapes. To analyze 4D trees, we need to jointly register and align all the 3D instances within a 4D tree. We perform this sequentially: each 3D tree-shape within a sequence is aligned to the next 3D tree-shape in the sequence, using \cref{eq:registration}. With this formulation, computing a geodesic, or the optimal deformation that aligns  $\srvfthreeDTree_2$
onto  $\srvfthreeDTree_1$ is straighforward. Let $\srvfthreeDTree\optimal_2 = (\srvfthreeDTree_2, \rotation\optimal,   \diffeotree\optimal, \permutetree\optimal)$. Since the metric is a weighted norm of $\ltwo$ distances, the geodesic $\geodesic\optimal$ between $\srvfthreeDTree_1$ and $\srvfthreeDTree_2$ is the straight line that connects $\srvfthreeDTree_1$ to $\srvfthreeDTree\optimal_2$, \ie $\geodesic\optimal(t) = (1 - t)  \srvfthreeDTree_1 + t \srvfthreeDTree_2, t\in [0, 1]$. For visualization, we map  $\geodesic\optimal(t)$ back to the non-linear space of 3D tree-shaped structures $\preshapespaceThreeDTreees$ using the inverse SRVF mapping, which has a closed analytical form; see~\cite{srivastava2010shape}.

%% 4D tree space
\section{The proposed space of 4D trees-shaped structures}
\label{sec:4D_tree_shape_space}
With this setup, a 4D tree-like shape becomes a time-parameterized 1D curve, or trajectory,   $\fourDcurve: [0, 1] \to \preshapesrvts$.
Thus,  analyzing  4D tree-shaped structures becomes the problem  of analyzing  1D curves in $ \preshapesrvts$. Here, we mathematically define the shape space $\preshapespaceFourDcurve$ of such curves and equip it with a proper metric (\cref{sect:4D tree shape space})  that will allow us to temporally register and compare such trajectories (\cref{sect:temopral_reg}).

\subsection{The shape space of 4D tree-shaped structures}
\label{sect:4D tree shape space}

Let $\fourDcurve: [0, 1] \to \preshapesrvts$ be a curve in $\preshapesrvts$ representing the SRVFT of a 4D tree-shaped object. Let $\preshapespaceFourDcurve$ be the space of all such curves. To temporally register, compare, and summarize samples of such curves, we need to define an appropriate metric on $\preshapespaceFourDcurve$ that is invariant to the execution rate of the curves. Mathematically, variations in the execution rate correspond to time-warping and thus can be represented using diffeomorphisms  $\reparmcurve: [0, 1] \to [0, 1] $ that map the temporal domain to itself. Two curves $\fourDcurve$ and $\fourDcurve \circ \reparmcurve $ only differ in their execution rates and thus should be deemed equivalent.  With this, the temporal registration of two 4D   trees  $\fourDcurve_1$ and $\fourDcurve_2$ becomes the problem of finding the optimal time warping $\reparmcurve^*$ that brings the two curves as close as possible to each other, with closeness measured using a metric $d(\cdot, \cdot)$:
$
    \reparmcurve^* = \argmin_{\reparmcurve\in\reparmcurvespace} d(\fourDcurve_1, \fourDcurve_2 \circ \reparmcurve),
%    \label{eq:distance_trajectories}
$ where  $\reparmcurvespace$ is the space of all diffeomorphisms $\reparmcurve: [0, 1] \to [0, 1] $. 
We are left with the problem of defining the metric, or measure of closeness, $d(\cdot, \cdot)$. As  time-warping of trajectories corresponds to the elasticity of curves, we follow the same approach used to compare branches (\cref{sect:srvf_3Dtree,eq:SRVF}), \ie instead of using a complex non-linear metric to measure the dissimilarity between two trajectories $\fourDcurve_1$ and $\fourDcurve_2$, we first map them to their SRVF space,  denoted by $\srvfbranch_1$ and $\srvfbranch_2$, and perform the analysis there. Working in the SRVF space has many benefits. \textbf{(First)}, the elastic metric in the original space reduces to the $\ltwo$ metric in the SRVF space. \textbf{Second}, under the $\ltwo$ metric, the action of the diffeomorphism  group $\reparmcurvespace$ is by isometries, \ie $\|\srvfbranch_1, \srvfbranch_2\| = \|\srvfbranch_1 \circ \reparmcurve,\srvfbranch_2\circ \reparmcurve\|$. \textbf{Third}, the SRVF is invertible, up to translation. Thus, one can perform all the analysis tasks in the SRVF space, which has an $\ltwo$ structure, and then map the results back to the original space for visualization without loss of information. Thus, temporal registration becomes:
\begin{equation}
    \reparmcurve^* = \argmin_{\reparmcurve\in\reparmcurvespace} \|\srvfbranch_1,\srvfbranch_2\circ \reparmcurve\|^2 \text{ and } d(\srvfbranch_1, \srvfbranch_2) = \inf_{\reparmcurve\in\reparmcurvespace} \|\srvfbranch_1, \srvfbranch_2 \circ \reparmcurve\|^2.
    \label{eq:srvf_distance_trajectories_registration}
\end{equation}

\noi Here,  $d(\cdot, \cdot) $ defines the rate-invariant distance between two trajectories (and thus two 4D shapes). Directly working with \cref{eq:srvf_distance_trajectories_registration} to solve the temporal registration problem is computationally very expensive since  $\fourDcurve$, which is a function, is of infinite dimension and when discretized, it will be of a very high dimension. Thus, we propose to learn a low dimensional subspace $\pcaspace$ of 3D trees, similar to PCA but on $\preshapespace_\srvfthreeDTree$. Then, we can treat 4D tree-shapes  as trajectories in $\pcaspace$, instead of the original space $\preshapespace_\srvfthreeDTree$, and perform their registration (\cref{sect:temopral_reg}) and statistical analysis (\cref{sec:statistical_analysis}) using the elastic metric of \cref{eq:srvf_distance_trajectories_registration}.

% \subsection{The subspace of 3D trees}
% \label{sec:pca_subspace}
The fact that  $ \preshapesrvts $ is equipped with \textbf{(1)} a proper metric that measures bending, stretching, and topological changes, and  \textbf{(2)} a mechanism for computing geodesics (\cref{sec:3D_shape_space}) allows us to learn a subspace, of finite dimension, that captures variability in collections of tree-shaped objects, similar to PCA in standard Euclidean spaces. Let $\{\threeDTree_i\}_{i =1}^{\totalthreeDtree}$ be a set of tree-shaped 3D objects and  $\{\srvfthreeDTree_i\}$ their corresponding SRVFTs. To compute the mean tree $\meantree$, we first compute the mean SRVFT  $\meansrvft $ and then map it back to the space of tree shapes. 
Mathematically, $\meansrvft $ is  the point in $\preshapesrvts$    that is as close as possible to all tree shapes in $\{\srvfthreeDTree_i\}_{i =1}^{\totalthreeDtree}$. The closeness is defined with respect to the metric of \cref{eq:invariantdistance}. In other words,
\begin{equation}
\small{
	\meansrvft = \argmin_{(\rotation_i, \diffeotree_i, \permutetree_i)_{ i=1}^\totalthreeDtree}
			\sum_{i=1}^{\totalthreeDtree}d^2_{\preshapesrvts}(\srvfthreeDTree, (\srvfthreeDTree_i, \rotation_i, \reparm_i, \permutetree_i)).
}
\label{eq:karcher_mean}
\end{equation}

\noi Solving  \cref{eq:karcher_mean} involves finding  $\meansrvft $,  known as the Karcher mean, while  registering every $\srvfthreeDTree_i$ onto $\meansrvft $. This can be efficiently done via a gradient descent approach:
\begin{enumerate}
	\item Set $\meansrvft =  \srvfthreeDTree_1$. 
	\item \label{step:for}For $i=1:\totalthreeDtree$,
		%\begin{itemize}
%			\item 
   Find $(\tilde\rotation_i,   \tilde\diffeotree_i, \tilde\permutetree_i)$ that optimally register  $\srvfthreeDTree_i$ onto $\meansrvft$ (\cref{eq:registration}). 
		%	\item Let $(\tilde\rotation_i,   \tilde\diffeotree_i, \tilde\permutetree_i)$ be the solution to \cref{eq:registration}.
%		\end{itemize}
	\item \label{set:assign} Set $\meansrvft =  \frac{1}{\totalthreeDtree}\sum_{i=1}^{\totalthreeDtree}(\srvfthreeDTree_i, \tilde\rotation_i, \tilde\diffeotree_i, \tilde\permutetree_i)$.
	\item Repeat steps~\ref{step:for} and~\ref{set:assign} until convergence, and return $\meansrvft$ and $(\tilde\rotation_i,   \tilde\diffeotree_i, \tilde\permutetree_i)_{i=1}^{\totalthreeDtree}$. 
\end{enumerate}

\begin{figure}[tb]
    \centering
    \includegraphics[trim={0cm 1cm 0cm 0cm},width=.9\linewidth]{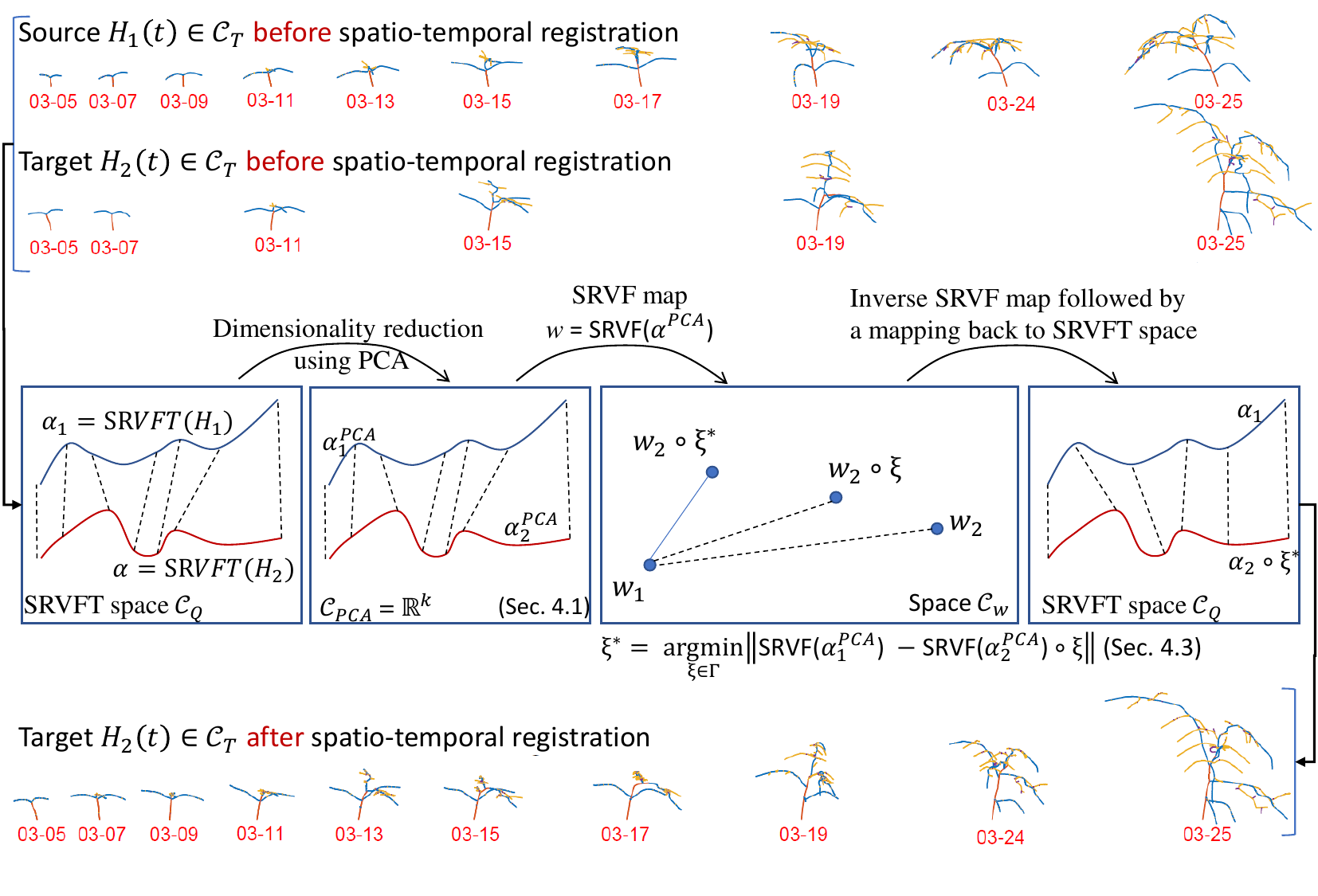} %{representation/SRVF_rep.pdf}
    \caption{The proposed framework for the analysis of 4D tree-shaped structures. The key idea is to represent 4D trees  $\fourDTree_1$ and $\fourDTree_2$  as curves in the   SRVFT space $\preshapesrvts$, which is Euclidean but of infinite dimension. By learning a PCA subspace $\pcaspace$, 4D tree-shaped structures become curves in $\real^k$. However, instead of using the nonlinear elastic metric in $\preshapespaceThreeDTreees$ to model temporal variability, we further map the curves to the SRVF space where the $\ltwo$ metric is equivalent to the full elastic metric. All the analysis can be performed in this space using standard vector calculus and mapped back to the original space for visualization. The computation time of the proposed temporal registration between two 4D plants is on average $0.037$s for tomato and $0.006$s for maize 4D plants.}
    \label{fig:all_mapping}
\end{figure}

\noi One can map $\meansrvft$ 
 back to the original tree shape to obtain the mean tree $\meantree$, which can be used for visualization. Let $\srvfthreeDTree_i = (\srvfthreeDTree_i,  \tilde\rotation_i, \tilde\diffeotree_i, \tilde\permutetree_i)$ be the SRVFT representation of $\srvfthreeDTree_i$ but optimally registered onto  the mean $\meansrvft$, and  $\bm{v}_i = \srvfthreeDTree_i - \meansrvft$.  
The  leading eigenvectors  $\Lambda_i$  of the  covariance matrix   $\displaystyle \covMatrix = \frac{1}{\totalthreeDtree-1}\sum_{i=1}^\totalthreeDtree\bm{v}_i\bm{v}_i^\top$ define the principal directions of variations while the corresponding eigenvalues $\lambda_i$ define the variance along the $i-$th eigenvector.  Thus, each SRVFT  $\srvfthreeDTree$ can be modeled as a  linear combination of the $k$ leading eigenvectors: $
	\srvfthreeDTree = \meansrvft + \sum_{i=1}^k a_i\sqrt{\lambda_i} \Lambda_i.
	%\label{eq:eigenprojection}
$
The coefficients $a_i\in \real$ are obtained by projecting $\srvfthreeDTree$ onto  eigenvectors $\Lambda_i$.   Thus, every tree shape $\srvfthreeDTree $ in the SRVFT space can now be represented as a point $p = (a_1, \dots, a_k)^\top$ in $\preshapesrvts = \real^k$, which is Euclidean and has a finite dimension.

\subsection{Temporal registration and geodesics between 4D trees}
\label{sect:temopral_reg}
With this representation, a 4D tree-shaped structure $\fourDcurve: [0, 1] \to \preshapesrvts$ becomes a trajectory  $\fourDcurveinpca$ 
in $\pcaspace = \real^k$. Thus,  the analysis framework presented in \cref{sect:4D tree shape space} for the pre-tree shape space can now be performed on $\real^k$. \cref{fig:all_mapping} summarizes the entire process. To temporally register two trajectories $\fourDcurveinpcaone$ and $\fourDcurveinpcatwo$, we first represent them using their SRVFs and then formulate the temporal registration problem as the one of finding an optimal 
diffeomorphism $\reparmcurve: [0, 1] \to [0, 1] $ that is a solution to   \cref{eq:srvf_distance_trajectories_registration}. We initialize $\reparmcurve$ to be the identity diffeomorphism: $\reparmcurve: [0, 1] \to [0, 1]$ such that $\reparmcurve(t) = t$. %In the implementation, we discretize the source domain $[0, 1]$ into $10$ equidistant samples. Thus, the discretized $\reparmcurve$ maps $\{0, 0.1, 0.2, 0.3, \cdots, 1.0\}$  to $\{0, 0.1, 0.2, 0.3, \cdots, 1.0\}$. 
During the optimization of \cref{eq:srvf_distance_trajectories_registration}, only the target domain is updated. In other words, a point $t$ in $[0, 1]$ will be mapped into another point $u$ in $[0, 1]$ such that \cref{eq:srvf_distance_trajectories_registration} is minimized. Since $u \in [0, 1]$, the search for the optimal match is over the entire continuous domain. Also, $\reparmcurve$ is enforced to be a diffeomorphism, \ie  $\reparmcurve(t_1) = u_1, \xi(t_2) = u_2$, and $t_1 < t_2$, then $u_1 < u_2$. 
The  spatio-temporal registration process is as follows;
%% Write the entire algorithm here
\begin{itemize}
    \item Let $S =\{\fourDTree_i\}_{i=1}^{\totalfourDtree}$ be a set of $\totalfourDtree$ 4D tree-shaped structures where $\fourDTree_i(t) \in \preshapespaceThreeDTreees$, for $t \in [0, 1]$, is a 3D tree-structured shape after normalization for scale. 
    
    \item \textbf{Step 1: Spatial registration.}
        \begin{itemize}
            \item Map every 4D tree $\fourDTree_i \in S$ to its SRVFT representation $\fourDcurve_i$. Thus, we obtain a new set $\{\fourDcurve_i\}_{i=1}^{\totalfourDtree}$ such that $\fourDcurve_i(t) = \text{SRVFT}(\fourDTree_i(t))$.
            \item Within-sequence  registration (Stage 1 of the spatial registration, \cref{sect:spatial_registraion_within}):
                \begin{itemize}
                    \item For $i=1$ to $\totalfourDtree$, spatially register every 3D tree-shape $\fourDcurve_i(t)$ in the $i-$th sequence $\fourDcurve_i$ to its next 3D tree-shape in the sequence (\cref{sect:spatial_registraion_within}). 

                    \item For simplicity of notation, let from now on  $\{\fourDcurve_i\}$ denote the new set. 
                \end{itemize}

            \item Cross sequence registration (Stage 2 of the spatial registration): to spatially align $\fourDcurve_1$ onto $\fourDcurve_2$,
                \begin{itemize}
                    \item Map $\fourDcurve_i, i \in \{1, 2\}$ to the PCA space $\pcaspace = \real^k$  to  obtain   $\{ \fourDcurveinpca_i \}$.% (\cref{sec:pca_subspace}). 

                    \item Interpolate, linearly, the samples of $\fourDcurveinpca_1$ and $\fourDcurveinpca_2$ and then discretize them at equidistances.

                    \item $\forall t$, spatially register  $\fourDcurveinpca_1(t)$ onto  $\fourDcurveinpca_2(t)$ (\cref{sect:spatial_registraion_within}).
                \end{itemize}
        \end{itemize}

    \item \textbf{Step 2: Temporal registration.} 
        \begin{itemize}
            \item Map the spatially registered  $\fourDcurveinpca_i$  to their  SRVF representation $\pcasrvf_i$. Let $\srvfpcaspace$ be the space of such SRVFs. For  any two curves $\pcasrvf_1$ and $\pcasrvf_2$:
            \begin{itemize}
                \item Find $\reparmcurve\optimal$ that optimally aligns $\pcasrvf_2$ onto $\pcasrvf_1$\\ by solving $
                \reparmcurve^* = \argmin_{\reparmcurve\in\reparmcurvespace} \|\pcasrvf_1,\pcasrvf_2\circ \reparmcurve\|^2$.

                \item Set $\pcasrvf_2 \leftarrow  \pcasrvf_2 \circ \reparmcurve\optimal$ and map it back to  $\preshapespaceThreeDTreees$  for visualization.
            \end{itemize}
        
        \end{itemize}

\end{itemize}

%\vspace{3pt}
\noindent
\textbf{4D geodesics:} The  advantage of the proposed representation is that the $\ltwo$ metric in the SRVF space $\srvfpcaspace $ of curves is equivalent to the full elastic metric. Thus, geodesics under the complex  metric become straight lines in this space, \ie the geodesic  $\geodbetnfourDtrees_\pcasrvf$ between $\pcasrvf$ and $\Tilde{\pcasrvf}_2$ is the linear interpolation:$
    \geodbetnfourDtrees_\pcasrvf(\tau)=(1-\tau)\pcasrvf+\tau\Tilde{\pcasrvf}_2,\ \tau\in[0,1]
$. 
To visualize it, we map all 4D trees in $\geodbetnfourDtrees_\pcasrvf$ to the original space of trees. \cref{fig:geod_after_reg} shows an example of a geodesic between two 4D trees  where each row corresponds to one 4D tree along the geodesic. The first row corresponds to $\geodbetnfourDtrees_\pcasrvf(0)= \pcasrvf_1$ and the last row corresponds to $\geodbetnfourDtrees_\pcasrvf(1)= \Tilde{\pcasrvf}_2$. The middle row, \ie  $\tau=0.5$,  is the mean 4D tree shape between $\pcasrvf_1$ and $\Tilde{\pcasrvf}_2$.

%% Summary statsitcs
\section{Statistical Analysis of 4D Tree Shapes}
\label{sec:statistical_analysis}
The ability to compute spatio-temporal correspondences and geodesics between 4D tree-shaped structures enables a wide range of 4D shape analysis tasks. In this section, we show how these fundamental tools can be used to compute a 4D atlas, which includes the mean and modes of variability,  of a collection of tree-shaped 4D structures.  Let $\{\fourDTree_1, \dots, \fourDTree_\totalfourDtree\}$ be a set of $\totalfourDtree$  4D tree shapes and $\{\pcasrvf_1, \dots, \pcasrvf_\totalfourDtree\}$ their corresponding trajectories in $\srvfpcaspace$. To simplify the notation, we assume that these 4D shapes are all spatially registered using the procedure described in \cref{sect:spatial_registraion_within} and \cref{sect:temopral_reg}. Thus, one can compute the mean, the modes of variations, and fit generative models for 4D tree-structured shape generation.

%With all the sample curves in $\preshapespacesrvffourDTree$, we compute statistics in that SRVF space of curves. 

\vspace{3pt}
\noi\textbf{Mean of 4D tree-like shapes.} It  is the 4D tree that is as close as possible to all of the 4D trees in the dataset under the specified metric. This is the Karcher mean  $\meansrvft $ and is mathematically defined as $
    \Bar{\pcasrvf}= \argmin_{\pcasrvf} \sum_{i=1}^{\totalfourDtree} \min_{\reparmcurve_i\in\reparmcurvespace} \| \pcasrvf - \srvffourDTree_i \circ \pcasrvf_i \|^2 
$, which we solve via a gradient descent approach:
\begin{enumerate}
	\item Set $\Bar{\pcasrvf} =  \pcasrvf_1$. 
	\item \label{step:for2}For $i=1:\totalfourDtree$
		\begin{itemize}
			\item Optimally register  $\pcasrvf_i$ onto $\Bar{\pcasrvf}$ using Step 2 of the algorithm of \cref{sect:temopral_reg}. 
			\item Let $\reparmcurve_i$ be the reparameterization that temporally aligns $\pcasrvf_i$ onto $\pcasrvf$.
		\end{itemize}
	\item \label{set:assign2} Set $\pcasrvf =  \frac{1}{\totalfourDtree}\sum_{i=1}^{\totalfourDtree} \pcasrvf_i \circ  \reparmcurve_i $.
	\item Repeat steps~\ref{step:for2} and~\ref{set:assign2} until convergence, and return 
	  $\Bar{\pcasrvf}$,  $\{\reparmcurve_i, \Tilde{\pcasrvf}_i = \pcasrvf_i \circ \reparmcurve_i\}_{i=1}^ \totalfourDtree$. 
\end{enumerate}

\noi Finally, the mean curve $\Bar{\pcasrvf}$ can be mapped back to a 4D tree, $\Bar{\fourDTree}$, which  represents the mean of the 4D tree-shaped structures in the dataset. The inverse mapping is performed following the procedure presented in \cref{sect:temopral_reg}. 

%This mean reveals the average deformation pattern of 4D tree-like shapes among a group in the temporal domain by computing over all the 4D tree shapes in the dataset.  

\vspace{3pt}
\noindent
\textbf{Modes of variations of 4D tree-shaped structures.} Since the SRVF space is Euclidean, we compute the principal directions of variation using linear PCA. Let $\{\Tilde{\pcasrvf}_i\}_{i= 1}^{\totalfourDtree}$ be the spatio-temporally registered 4D tree structures in  $\srvfpcaspace$. We first compute their covariance matrix and then take its $k$-leading eigenvalues $u_i, i=1, \dots,k$ and their corresponding eigenvectors $\eigenvectorcurvepca_i, i=1, \dots, k$. This allows us to represent any curve  (\ie 4D trees) in the $i^{th}$ principal direction as $\pcasrvf_\tau=\Bar{\pcasrvf} + \tau\sqrt{u_i}\eigenvectorcurvepca_i,\ \tau\in\real$. For visualization, we map $\pcasrvf_\tau$ back to the original representation of 4D tree shapes, using the one-to-one inversion  discussed in \cref{sect:temopral_reg}.

\vspace{3pt}
\noindent
\textbf{4D tree-like shape generation:} Now that we have the mean $\Bar{\pcasrvf}$ and $k$ principal directions of variation, $\eigenvectorcurvepca_i, i=1,\dots, k$, we can generate  a 4D tree-like shape by  sampling $k$ real values $\tau_i\in\real$ and then computing $ 
    \pcasrvf= \Bar{\pcasrvf} + \sum_{i=1}^k\tau_i\sqrt{u}_i\eigenvectorcurvepca_i, \ \tau_i\in\real.
 %   \label{eq:generation}
$
Next, $\pcasrvf$ can be mapped back to the original tree-shape space for visualization using the one-to-one inversion discussed in \cref{sect:temopral_reg}. Note that this generation process is random since the real values $\tau_i$  are randomly picked. However, one can control the generation process, \eg by restricting these values to be within a specific range to ensure the plausibility of the generated samples.

%% Results
\begin{table}[t]%{r}{\textwidth}
\centering
    \caption{Statistics of the error between pairs of source and target 4D plants before spatial registration and after registration using our method, \cite{chebrolu2021registration}, and \cite{skeletal_extraction}. }
    \label{tab:spatial_reg_error}

    \begin{tabular}{|@{ }c|c|c|c|c|c|c|c|c|c|c|c|c@{ }|}
        \hline
             & \multicolumn{3}{|c|}{\textbf{Before reg.}}  &  \multicolumn{3}{|c|}{\textbf{After reg. (ours)}} & \multicolumn{3}{|c|}{\textbf{After reg.~\cite{chebrolu2021registration}}} & \multicolumn{3}{|c|}{\textbf{After reg.~\cite{skeletal_extraction}}}\\
        \cline{2-13}
             & Mean & Median & Std. & Mean & Median & Std. & Mean & Median & Std. & Mean & Median & Std. \\
        \hline
        Tomato &  $325$ & $ 347$ & $52$ & $ \bf{136}$ & $ \bf{136}$ & $ \bf{27} $  &$416$  & $430$ & $85$ & $261$ & $245$ & $56$  \\
        \hline
        Maize &  $ 35 $ & $ 30 $ &$ 8$ &$\bf{10}$ & $\bf{9}$ &$\bf{2.5}$ &  $554$ & $496$ & $216$ & $19$ & $18$ & $5$\\       
        \hline
    \end{tabular}
\end{table}

\section{Results and discussions}
\label{sec:result}
We demonstrate the proposed framework and evaluate its performance on the Pheno4D dataset \cite{pheno4d}, which contains 4D models of seven tomato and seven maize plants. All the models are not registered onto each other, neither spatially nor temporally; see the Supplementary Material for further details and results with ablation studies. %, including a detailed analysis of the computation time.

\subsection{Spatiotemporal Registration}
\label{sect:registration}

\noindent
\textbf{(1) Spatial registration.} \cref{fig:spatial registration}  shows an example of the spatial registration of four complex 3D tree-shaped structures of  tomato plants. For clarity, we only show the branch-wise correspondences up to the second layer. 
Figs. 2, 3, and 4 in the Supplementary Material show, respectively, an example of the spatial registration within and across two 4D tomato plants as well as a zoom-in onto individual trees.  This result shows that our framework is efficient in finding correct branch-wise correspondences between complex tree-shaped structures. We evaluate the quality of the proposed spatial registration method and compare it with state-of-the-art techniques such as \cite{chebrolu2021registration,skeletal_extraction} on eight pairs of 4D tomato plants and nine pairs of  4D maize plants. We use three quantitative measures:

\begin{wraptable}{r}{0.55\textwidth}
\centering
    \caption{Cycle consistency errors $(\downarrow)$. }
    \label{tab:cycle_consistancy}
   \resizebox{\linewidth}{!}{
    \begin{tabular}{|@{ }c|c|c|c|c|c|c@{ }|}
        \hline
             & \multicolumn{3}{|c|}{\textbf{Our method}}  &  \multicolumn{3}{|c|}{Pan \etal~\cite{skeletal_extraction}} \\
        \cline{2-7}
          $\epsilon$   & Mean & Median & Std. & Mean & Median & Std. \\
        \hline
        $0.1$ & $\bf{0.94\%}$ & $\bf{0.08\%}$  & $\bf{2.01\%}$  & $10.33\%$ & $7.09\%$ & $10.6\%$  \\
        \hline
        $0.05$ & $\bf{2.00\%}$  & $\bf{0.15\%}$  &  $\bf{3.62\%}$ & $15.89\%$ & $17.88\%$ & $10.79\%$  \\  
        \hline
        $0.02$ & $\bf{2.87\%}$  &  $\bf{0.65\%}$ &  $\bf{4.17\%}$ & $20.39\%$ & $22.22\%$ & $9.83\%$  \\
        \hline
        $0.01$ & $\bf{5.45\%}$ & $\bf{2.76\%}$ &  $\bf{6.15\%}$ & $21.95\%$ & $22.22\%$  &  $8.64\%$ \\
        \hline
    \end{tabular}
    }
\end{wraptable}

%\vspace{1pt}
\noi \textit{- Geodesic length between 4D tree-shapes}. \cref{tab:spatial_reg_error} reports the mean, median, and standard deviation of the geodesic distances before and after spatial registration of our method. We can see that the geodesic distance becomes much smaller after spatial registration with our method, compared to~\cite{chebrolu2021registration} and \cite{skeletal_extraction} as branches get correctly aligned across 3D trees. The residual error is due to differences in structure and growth rates between 4D plants.

\begin{wraptable}{r}{0.5\textwidth}
\centering
 \vspace{-30pt}
\caption{Statistics of the error between pairs of 4D source and target plants before and after their temporal registration.}
\label{tab:temporal_reg_error}
\resizebox{\textwidth}{!}{%
\begin{tabular}{|ccccccc|}
    %\hline
    \multicolumn{7}{c}{\textbf{(a)} Between a source and a target 4D tree.}\\ 
    \hline

    & 
    \multicolumn{3}{|c|}{\textbf{Before reg.}} &
    \multicolumn{3}{c|}{\textbf{After reg.}} \\ 
    \cline{2-7}
    &
    \multicolumn{1}{|c|}{Mean} &
    \multicolumn{1}{c|}{Median}&
    \multicolumn{1}{c|}{Std.} &
     \multicolumn{1}{c|}{Mean} &
     \multicolumn{1}{c|}{Median} &
    Std. \\ 
    \cline{1-7}

  \multicolumn{1}{|r|}{Tomato} &
  \multicolumn{1}{c|}{$382$} &
  \multicolumn{1}{c|}{$384$}&
  \multicolumn{1}{c|}{$69$} &
  \multicolumn{1}{c|}{$\textbf{279}$} &
  \multicolumn{1}{c|}{$\textbf{282}$} &
  \multicolumn{1}{c|}{ $\textbf{47}$}\\ 
   \cline{1-7} 

\multicolumn{1}{|r|}{Maize} &
  \multicolumn{1}{c|}{$59$} &
  \multicolumn{1}{c|}{$60$} &
  \multicolumn{1}{c|}{$12$} &
  \multicolumn{1}{c|}{$\textbf{34}$} &
  \multicolumn{1}{c|}{$\textbf{37}$} &
  $\textbf{11}$ \\
  
  \hline
  \multicolumn{7}{c}{\textbf{(b)} Between a 4D tree and its randomly} \\ 
  \multicolumn{7}{c}{warped version.} \\ 
  \hline
     &
  \multicolumn{3}{|c|}{\textbf{Before reg.}} &
  \multicolumn{3}{c|}{\textbf{After reg.}} \\ 
  \cline{2-7} 

  &
  \multicolumn{1}{|c|}{Mean} &
  \multicolumn{1}{c|}{Median} &
  \multicolumn{1}{c|}{Std.} &
  \multicolumn{1}{c|}{Mean} &
  \multicolumn{1}{c|}{Median} &
  Std. \\ 
  \cline{1-7} 

    \multicolumn{1}{|r|}{Tomato} &
  \multicolumn{1}{c|}{$356$} &
  \multicolumn{1}{c|}{$375$} &
  \multicolumn{1}{c|}{$138$} &
  \multicolumn{1}{c|}{$\textbf{26}$} &
  \multicolumn{1}{c|}{$\textbf{27}$} &
  $\textbf{5}$\\ 
   \cline{1-7} 
   
\multicolumn{1}{|r|}{Maize} &
  \multicolumn{1}{c|}{$71$} &
  \multicolumn{1}{c|}{$70$} &
  \multicolumn{1}{c|}{$8$} &
  \multicolumn{1}{c|}{$\textbf{15}$} &
  \multicolumn{1}{c|}{$\textbf{16}$} &
  $\textbf{3}$ \\ \hline
\end{tabular}%
}
\end{wraptable}

%\vspace{1pt}
\noi \textit{- Cycle consistency error}.
To demonstrate further the quality of our registration, we have measured the cycle consistency error of the spatial registration. Given a source and a target 3D trees, the registration maps a point $\textbf{x}$ on the source to a point $\textbf{y}$ on the target. We then map $\textbf{y}$, using the registration procedure, back onto the source tree to lead to a point $\textbf{x}'$. The registration procedure is accurate if  $\textbf{x}'$ and $\textbf{x}$ are very close to each other. Thus, we define the registration error as the percentage of points $\textbf{x}$ whose cycle consistency distance $\|\textbf{x} - \textbf{x}'\|$ is higher than a threshold $\epsilon$. \cref{tab:cycle_consistancy}  summarizes the mean, median, and standard deviation of this error for our method and \cite{skeletal_extraction} (the lower the error, the better) for different values of $\epsilon$. Our method significantly outperforms the state-of-the-art.

% \vspace{2pt}
% \noi\textit{- Description length} defined as the number of eigenvectors needed to characterize $x\%$ of the variability within the dataset.  $x$ is referred to as the cumulative energy. From Fig. 5 %\cref{I-fig:comparison} 
% in the Supplementary Material, we can see that our method shows stable preservation of energy. Although \cite{skeletal_extraction} and \cite{chebrolu2021registration} preserve equal energy to ours, these methods show a higher drop when using less than $65\%$ of the eigenvectors.

\vspace{1pt}
\noi\textit{- Computation time.}  On average,~\cite{chebrolu2021registration} requires $30$s,~\cite{skeletal_extraction} requires $20.1$s, while our method requires only $2.3$s to register two 3D trees. This shows that the proposed method is significantly faster than \cite{chebrolu2021registration} and \cite{skeletal_extraction}.

\vspace{1pt}
\noindent
\textbf{(2) Temporal registration.}  To demonstrate the proposed temporal registration, we first randomly resample the 4D sequences to  simulate unsynchronised 4D plants and then take pairs of these re-sampled 4D sequences and realign them, temporally, using the proposed temporal registration tools.  \cref{fig:all_mapping} shows an example of a temporal registration between complex 4D tree-shaped structures that grow at different rates. We can see that, after the temporal registration, the growth rate becomes closer to the groundtruth (see also Fig. 6 %\cref{I-fig:temporal_reg_result} 
of the Supplementary Material). The Supplementary Material also provides more visual results.  Similar to the spatial registration, we use the geodesic distance between 4D trees, before and after temporal registration, as a measure of the quality of registration (the smaller the better). \cref{tab:temporal_reg_error} reports the mean, median, and standard deviation of those errors for three different cases. For all of the cases, our proposed temporal registration significantly reduces the geodesic distances between 4D plants: \textit{In \cref{tab:temporal_reg_error}-(a)}, we randomly choose six pairs of 4D tomato plants and four pairs of maize plants. Then, we compute errors between each pair before and after temporal registration. We observe that the proposed temporal registration makes the target  closer to the source  4D plant. Note that there is a residual error which is due to the differences in the branching structures. \textit{In \cref{tab:temporal_reg_error}-(b)}, we register  randomly warped 4D plants to their ground truth. Since both 4D plants have the same branching structures and are only temporarily warped, temporal registration significantly reduces the error, which demonstrates that the proposed temporal registration can align two 4D  shapes with minimal error.

Overall, the temporal registration takes $6$ to $37$ms; see Table 1 in the Supplementary Material, which also  provides an ablation study that analyzes the spatio-temporal registration of 4D tree-shaped objects.

\subsection{Geodesics between 4D Trees}
\label{sect:geodesic}
\begin{figure}[tb]
    \centering
    \includegraphics[width=.9\linewidth]{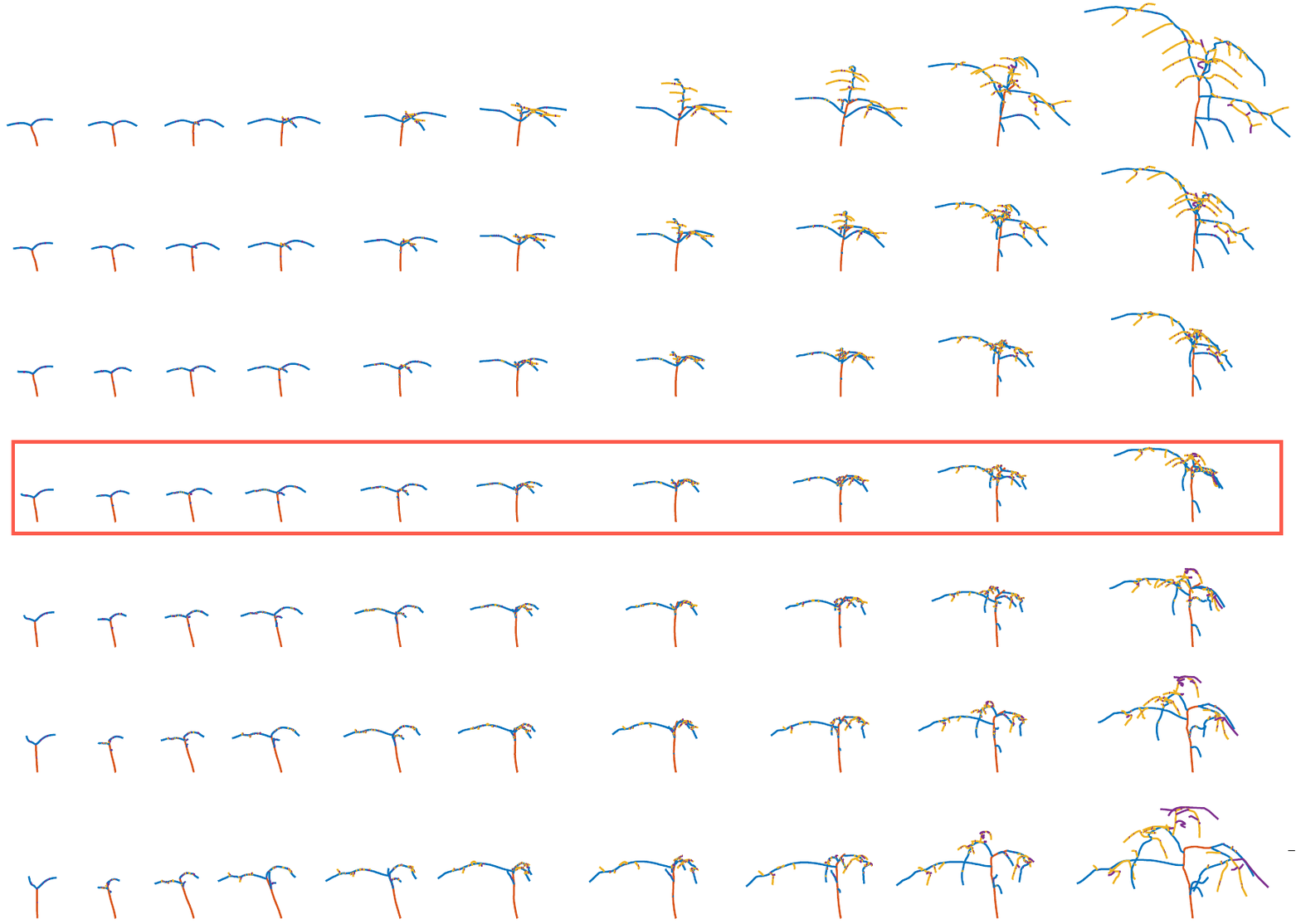}
    \caption{The 4D geodesic  between the two 4D tree shapes of Fig. 8 % \cref{I-fig:geod_before_reg}
    in the Supplementary Material after  spatiotemporal registration. The highlighted  row  is the mean 4D tree. This geodesic computation requires on average $0.025$s for 4D tomato plants.}
    \label{fig:geod_after_reg}
\end{figure}

%\cref{I-fig:geod_before_reg} 
Fig. 8 in the Supplementary Material shows a geodesic path, before spatiotemporal registration, between a source 4D plant (top row) and a target 4D plant (bottom row). The in-between rows correspond to  4D plants sampled at equidistances along the geodesic path between the source and target. In this example, the target has a different growth rate than the source. It also has missing samples. Thus, we interpolate it and sample it at the same time intervals as the source but we do not perform spatiotemporal registration. Consequently, the target follows a different growth pattern than the source. As we can see in this figure, since the two 4D plants are not spatio-temporally registered, the 4D geodesic does not look realistic as the intermediate 4D plants along the geodesic exhibit significant shrinkage. \cref{fig:geod_after_reg} shows the geodesic between the same 4D plants but after spatiotemporal registration using the proposed framework. We can clearly see that the target 4D plant (last row) is temporally well aligned with the source 4D plant (first row). Also, we can also see that the branches of the intermediate plants along the geodesic path  do not suffer from the shrinkage observed in Fig. 8 %\cref{I-fig:geod_before_reg}, 
in the Supplementary Material prior to the registration. %The Supplementary Material provides more results with ablation studies.

\begin{figure}[t]
    \centering
    \includegraphics[width=0.9\textwidth]{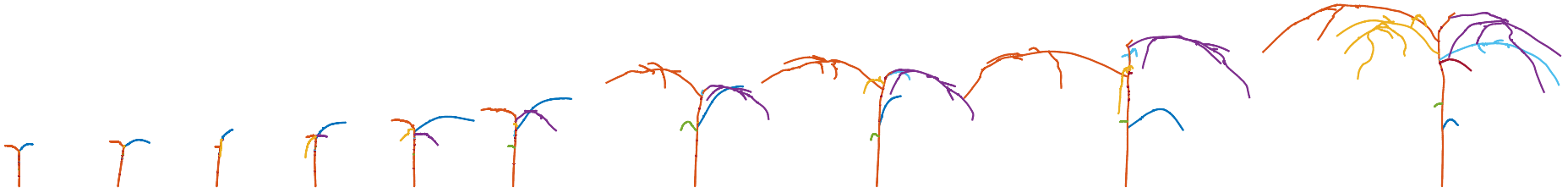}
    \caption{The mean 4D plant shape of the seven registered 4D tomato plants in Fig. 10-b
    %\cref{I-fig:Co-registration_b} 
    in the Supplementary Material. The computation time is in the order of $0.0006$s. %It has the same color code as the registered trees in \cref{fig:Co-registration_b}.
    }
    \label{fig:mean_shape}
\end{figure}

\begin{figure}[t]
    \centering
    %\begin{subfigure}{1\linewidth}
    %    \centering
        \includegraphics[width=.9\linewidth]{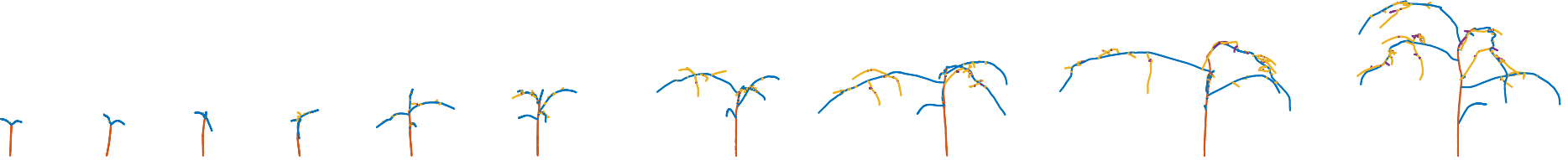}\\
    %\end{subfigure}
    %\begin{subfigure}{1\linewidth}
    %    \centering
        \includegraphics[width=.9\linewidth]{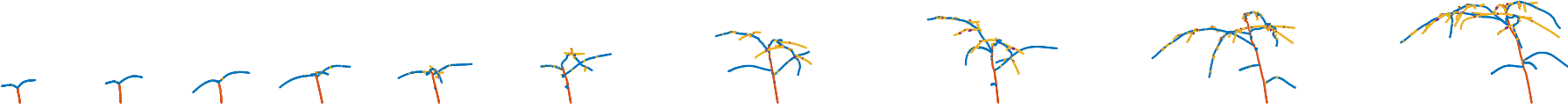}\\
    %\end{subfigure}
    %\begin{subfigure}{1\linewidth}
    %    \centering
        \includegraphics[width=.9\linewidth]{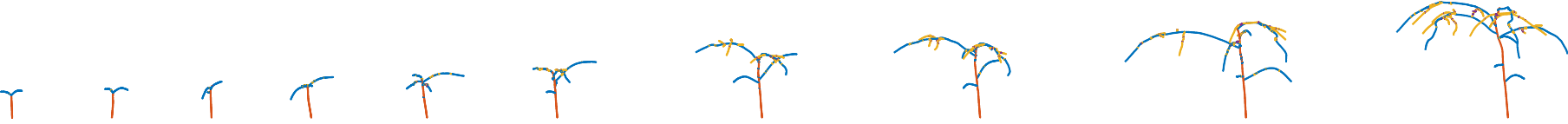}\\
    %\end{subfigure}
    %\begin{subfigure}{1\linewidth}
    %    \centering
        \includegraphics[width=.9\linewidth]{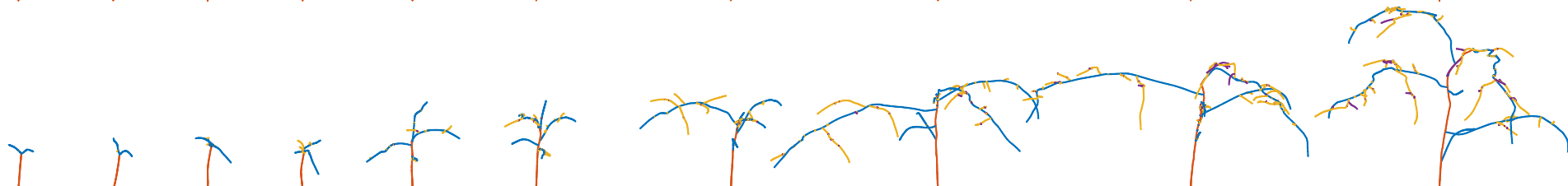} \\
    %\end{subfigure}
    \caption{Randomly generated 4D tomato plants (one 4D plant per row). The model requires on average $0.0001s$ to generate a 4D tomato plant.}
    \label{fig:rand_samples.pdf}
\end{figure}

\subsection{Summary statistics and 4D tree-shae synthesis and generation}
\label{sect:summary_statistics}
We took seven unregistered 4D tomato plants from the Pheno4D dataset and computed their mean and modes of variations using the proposed framework. % \cref{I-fig:Co-registration} 
Fig. 10 in the Supplementary Material shows the seven 4D tomato plants before (Fig. 10-a) %(\cref{I-fig:Co-registration_a}) 
and after (Fig. 10-b)
%\cref{I-fig:Co-registration_b}) 
co-registering them onto each other using the proposed spatiotemporal registration. \cref{fig:mean_shape}  shows the mean 4D plant of these seven 4D plants computed using the proposed framework. We can see that the mean captures the main characteristics of these seven 4D plants. %\cref{I-fig:modes} 
Fig. 12 in the Supplementary Material shows the first and second principal directions of variation that represent the spatiotemporal shape variability in the seven 4D tomato plants. The Supplementary Material provides more results.

%\subsection{4D Tree-shape synthesis and generation}
%\label{sect:synthesis}
\cref{fig:rand_samples.pdf} shows four randomly generated  4D tomato plants. In this experiment, the generative model is learned from seven 4D tomato plants. To obtain plausible random 4D plants, we restricted the randomization within $-3$ to $+3$ times the standard deviation along each principal direction of variation. The Supplementary Material provides additional examples of 4D tree-shaped structures randomly generated from the learned generative model.

%% Conclusion
\vspace{-12pt}
\section{Conclusion}
\label{sec:conclusion}
%\vspace{-6pt}
We have proposed a novel theoretical framework and a set of new computational tools for analyzing tree-shaped 4D structures, \ie 3D objects that have a tree structure and grow and deform over time. These tools include mechanisms for the spatiotemporal registration of and geodesics computation between tree-shaped 4D structures. These are the key building blocks for learning statistical models and generative models from collections of tree-shaped 4D structures such as growing plants. Our key contribution is to model the spatiotemporal variability in 4D tree shapes as trajectories in the SRVF space, reducing the problem to that of analyzing curves in an Euclidean space.  As demonstrated in the experiments, the proposed computational tools can handle complex tree-shaped 4D objects that undergo complex non-rigid and topological deformations. Although we evaluated the framework on 4D plants, it is general and can be used for other types of tree-shaped structures such as neuronal structures in the brain,  blood vessels, and airway trees.  The current framework is limited to the skeleton structure of tree-shaped 4D objects. In the future, we plan to extend the framework to handle the full 3D geometry of the branches in tree-shaped 3D and 4D objects.

\vspace{1pt}
\noi\textbf{Potential negative impact.} To the best of our knowledge, this work may not have any negative impact on society.

\clearpage
\section*{Acknowledgements}
This work is supported by the Australian Research Council (ARC) Discovery Projects no. DP220102197, DP210101682, and DP210102674. Tahmina Khanam is funded by the Murdoch University International Postgraduate Scholarship. Guan Wang is supported by the Scientific Research Foundation for Yangtze Delta Region Institute of University of Electronic Science and Technology of China (Huzhou) under Grant U032200114.  Anuj Srivastava is supported by NSF DMS 1953087 and  NSF DMS 241374. The authors  would like  to thank Adam Duncan for sharing with us the code of~\cite{duncan_bifurcation}. The source code of this project is publicly available, for research purposes, at \url{https://github.com/Tahmina979/4Dtreeshape_project}.

\bibliographystyle{splncs04}
\bibliography{main}
\end{document}